\documentclass[runningheads]{llncs}
\usepackage[figuresright]{rotating}
\usepackage{mathtools}
\usepackage{xcolor}  
\usepackage{graphicx}
\usepackage{bm}
\usepackage{makecell}
\usepackage{bbm}
\usepackage{xspace}
\usepackage{float}
\usepackage{color}
\usepackage{colortbl}
\usepackage{dsfont}
\usepackage{multirow}
\usepackage{multicol}
\usepackage{pgfplots}
\usepackage{algorithm}
\usepackage{algpseudocode}
\usepackage{adjustbox}
\usepackage{listings}
\usepackage{bbding}
\usepackage{pifont}
\usepackage{wasysym}
\usepackage{amssymb}
\usepackage{tabularx}
\usepackage[misc]{ifsym}

\begin{document}
\title{ICDAR 2023 Competition on Structured Text Extraction from Visually-Rich Document Images}

\author{Wenwen Yu\inst{1}\thanks{Contributed equally.} \and
Chengquan Zhang\inst{2}$^\star$ \and
Haoyu Cao\inst{3}$^\star$ \and
Wei Hua\inst{1} \and
Bohan Li\inst{2} \and
Huang Chen\inst{3} \and
Mingyu Liu\inst{1} \and
Mingrui Chen\inst{1} \and
Jianfeng Kuang\inst{1} \and
Mengjun Cheng\inst{5} \and
Yuning Du\inst{2} \and
Shikun Feng\inst{2} \and
Xiaoguang Hu\inst{2} \and
Pengyuan Lyu\inst{2} \and
Kun Yao\inst{2} \and
Yuechen Yu\inst{2} \and
Yuliang Liu\inst{1} \and
Wanxiang Che\inst{6} \and
Errui Ding\inst{2} \and
Cheng-Lin Liu\inst{7} \and
Jiebo Luo\inst{8} \and
Shuicheng Yan\inst{9} \and
Min Zhang\inst{6} \and
Dimosthenis Karatzas\inst{4} \and
Xing Sun\inst{3} \and
Jingdong Wang\inst{2} \and
Xiang Bai\inst{1}\textsuperscript{(\Letter)}%
}

\authorrunning{W. Yu et al.}
\institute{Huazhong University of Science and Technology\\ \email{\{wenwenyu, xbai\}@hust.edu.cn}\\\and
Baidu Inc.\\
\email{zhangchengquan@baidu.com} \and
Tencent YouTu Lab\\
\email{\{rechycao, huaangchen, winfredsun\}@tencent.com} \and
Universitat Autónoma de Barcelona \and
Peking University\and Harbin Institute of Technology\and CAS Institute of Automation\and University of Rochester\and Sea AI Lab}

\maketitle             

\begin{abstract}
Structured text extraction is one of the most valuable and challenging application directions in the field of Document AI. However, the scenarios of past benchmarks are limited, and the corresponding evaluation protocols usually focus on the submodules of the structured text extraction scheme. In order to eliminate these problems, we organized the ICDAR 2023 competition on Structured text extraction from Visually-Rich Document images (SVRD). We set up two tracks for SVRD including Track 1:~\textbf{HUST-CELL} and Track 2:~\textbf{Baidu-FEST}, where~\textbf{HUST-CELL} aims to evaluate the end-to-end performance of~\textbf{C}omplex~\textbf{E}ntity \textbf{L}inking and~\textbf{L}abeling, and~\textbf{Baidu-FEST} focuses on evaluating the performance and generalization of Zero-shot /~\textbf{Fe}w-shot ~\textbf{S}tructured \textbf{T}ext extraction from an end-to-end perspective. Compared to the current document benchmarks, our two tracks of competition benchmark enriches the scenarios greatly and contains more than 50 types of visually-rich document images (mainly from the actual enterprise applications). The competition opened on 30th December, 2022 and closed on 24th March, 2023. There are 35 participants and 91 valid submissions received for Track 1, and 15 participants and 26 valid submissions received for Track 2. In this report we will presents the motivation, competition datasets, task definition, evaluation protocol, and submission summaries. According to the performance of the submissions, we believe there is still a large gap on the expected information extraction performance for complex and zero-shot scenarios. It is hoped that this competition will attract many researchers in the field of CV and NLP, and bring some new thoughts to the field of Document AI.

\end{abstract}
\section{Introduction}
In recent years, the domain of Document AI has gradually become a hot research topic. As one of the most concerned Document AI technologies, structured text extraction aims to capture text fields with specified semantic attributes from complex visually-rich documents (VRDs). It is widely used in many applications and services, such as customs information inspection, accounting in the financial field, office automation, and so on.

In the past, several benchmarks have been established in the community, such as FUNSD~\cite{jaume2019funsd}, CORD~\cite{park2019cord}, XFUND~\cite{xu2022xfund}, EPHOIE~\cite{wang2021ephoie}, etc., to measure relative technical efforts. However, there are many imperfections in these benchmarks and the corresponding evaluation protocols. The obvious shortcomings are as follows: 1) The performance of structured text extraction is not evaluated from the end-to-end perspective, but is disassembled into three independent functional modules, namely text detection~\cite{liao2020real,zhou2017east,zhang2019look}, text recognition~\cite{shi2016end,yu2020towards,fang2021read} and entity labelling or linking)~\cite{xu2020layoutlm,appalaraju2021docformer,li2021structext,yu2021pick}, for respective evaluation, which is not intuitive for downstream applications. 2) The scenarios covered by the above benchmarks are relatively few, and only focus on a certain receipt or form scenario, which is difficult to guide objectively evaluate the effectiveness and robustness of the model.

Therefore, we propose a new structured text extraction competition benchmark including two tracks, which covers the most abundant visually-rich document images of scenarios and types as far as we know. The whole benchmark will contain more than 50 document types and more than 100 semantic attributes of text fields. In order to evaluate the performance of structured information extraction from an end-to-end perspective, we will set up two tracks: (1) end-to-end ~\textbf{C}omplex~\textbf{E}ntity \textbf{L}inking and~\textbf{L}abeling created by \textbf{H}uanzhong \textbf{U}niversity of \textbf{S}cience and \textbf{T}echnology~(\textbf{HUST-CELL}), where we have designed entity linking and labeling tasks. (2) end-to-end ~\textbf{Fe}w-shot ~\textbf{S}tructured \textbf{T}ext extraction created by \textbf{Baidu} Company (\textbf{Baidu-FEST}) to explore the generalization and robustness of the submitted models, where we have newly designed zero-shot and few-shot structured text extraction tasks. The motivation and relevance to ICDAR community including:
\begin{itemize}
\item The intelligent analysis of visually-rich document images has always been an important domain of concern for ICDAR community. Its core technologies include text detection, recognition and named-entity recognition. Our proposed competition is the first time to guide the evaluation of the effect and generalization of structured text extraction scheme from an end-to-end perspective, which is more valuable but more challenging.

\item This competition aims to further connect researchers from both the document image understanding and NLP communities to bring more inspiration.
\end{itemize}

\subsection{Competition Organization}
ICDAR 2023 competition on SVRD is organized by a joint team, including Huazhong University of Science and Technology, Baidu Inc., Tencent YouTu Lab, Universitat Autónoma de Barcelona, Peking University, Harbin Institute of Technology, CAS Institute of Automation, University of Rochester, and Sea AI Lab.

We organize the SVRD competition on the Robust Reading Competition (RRC) website~\footnote{https://rrc.cvc.uab.es/?ch=21}, where provide corresponding download links of the datasets, and user interfaces for participants and submission page for their results~\footnote{Scores achieved using the ChatGPT large model interface during the competition are temporarily excluded from the leaderboard.}. Great support has been received from the RRC web team.

\section{Related Works}
This section discusses most of the well-received visually-rich document benchmarks as following:

In 2019, the Robust Reading Competition (RRC) web portal introduced a new challenge, known as Scanned Receipts OCR and Information Extraction reading competition which also commonly named as SROIE~\cite{huang2019icdar2019}. The main feature of SROIE is that all the images are collected from scanned receipts. It contains of 626 receipts for training and 347 receipts for testing, and each receipt only contains four predefined values: company, date, address, and total.

Meanwhile, EATEN~\cite{guo2019eaten} constructs a dataset of 1,900 real images in train ticket scenarios, and it only has entity values annotation without OCR annotation. CORD~\cite{park2019cord} consists of 1,000 Indonesian receipts, which contains images and box/text annotations for OCR, and multi-level semantic labels for parsing, which can be used to address various OCR and entity extraction tasks.

FUNSD~\cite{jaume2019funsd} dataset was presented at the ICDAR workshop in 2019. FUNSD is a form understanding benchmark with 199 real, fully annotated, scanned form images, such as marketing, advertising, and scientific reports, which is split into 149 training samples and 50 testing samples. FUNSD dataset is suitable for a variety of tasks including text detection, recognition, entity labelling, etc. And its entity has only three types of semantic attributes including question, answer and header.

XFUND~\cite{xu2022xfund} is an extended version of FUNSD that was proposed in 2021. It is launched to introduce the multi-script structured text extraction problem. It consists of  human-labeled forms with key-value pairs in 7 languages (Chinese, Japanese, Spanish, French, Italian, German, Portuguese). Each language includes 199 forms, where the training set includes 149 forms, and the test set includes 50 forms.

EPHOIE~\cite{wang2021ephoie} was also released in 2021, and contains 1,494 images which are collected and scanned from real examination papers of various schools in China. There are 10 text entity types including Subject, Test Time, Name, School, Examination Number, Seat Number, Class, Student Number, Grade, and Score. 
As an overview, Table~\ref{table:Dataset_description} lists the details of above datasets.
\begin{table*}[htbp]
\caption{Existing Visually-rich Document Images Datasets.}
\label{table:Dataset_description}
\begin{center}
\setlength{\tabcolsep}{1mm}
{
\renewcommand{\arraystretch}{1.1}
\scalebox{0.88}{
\begin{tabular}{c c c c c c}
\hline
\textbf{Dataset} & \textbf{Image number(Train/Test)} & \textbf{Language} & \textbf{Granularity} & \textbf{Document type} &  \textbf{Year}\\
\hline
SROIE  & 973(626/347) & English & Word & Receipt &2019 \\
CORD  & 1,000(800/200) & English & Word & Receipt & 2019   \\
EATEN  & 1,900(1,500/400) & Chinese & w/o OCR & Train Ticket & 2019   \\
FUNSD  & 199(149/50) & English & Word &Form & 2019  \\
XFUND  & 1,393(1,043/350) & 7 languages & Word/Line & Form & 2021   \\
EPHOIE  & 1,494(1,183/311) & Chinese & Character & Paper & 2021   \\
\hline
\end{tabular}}}
\end{center}
\end{table*}

\section{Benchmark Description}
\subsection{Track 1: HUST-CELL}
\begin{figure*}[ht]
 \centering
 \includegraphics[width=0.8\textwidth]{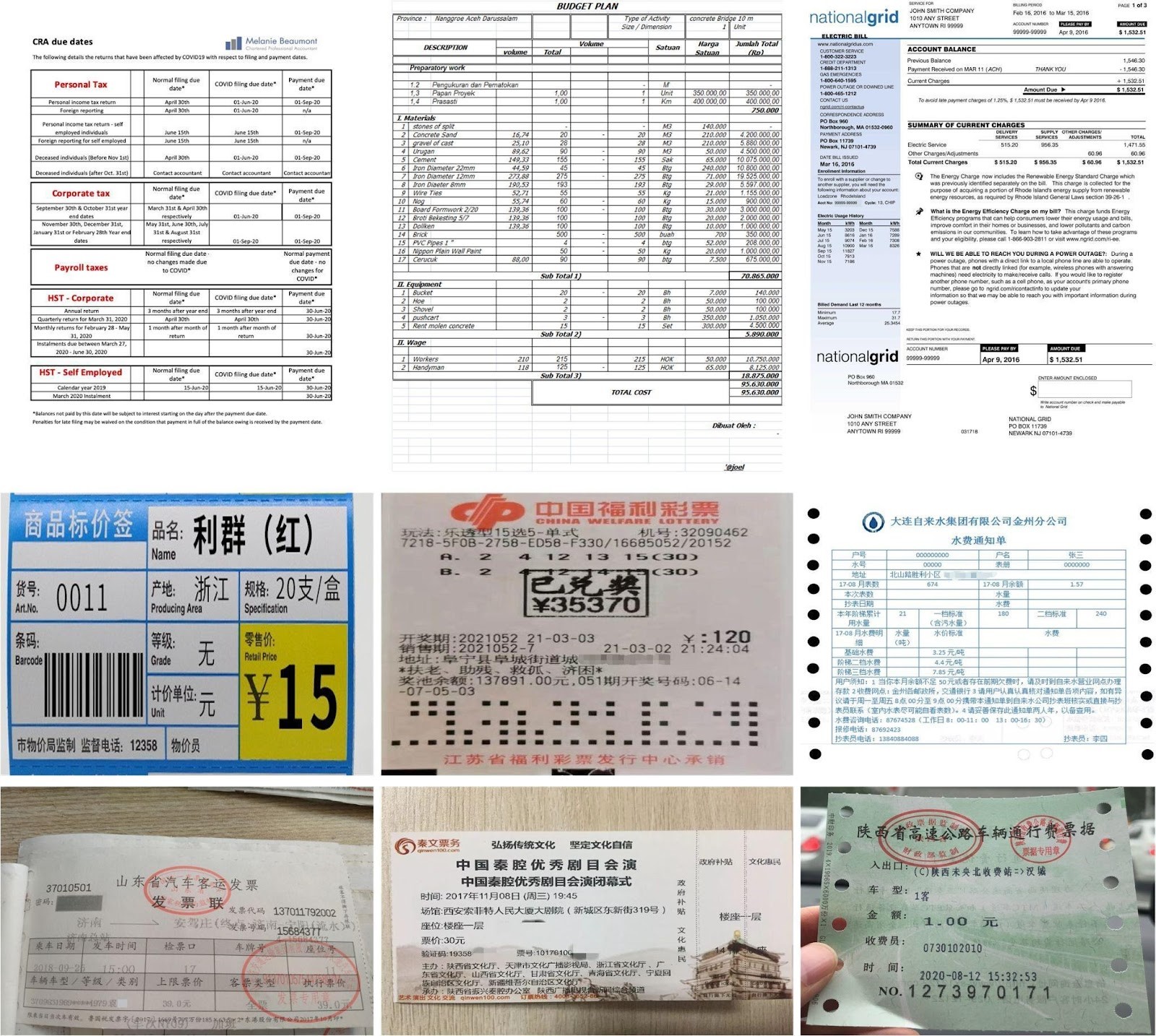}
\caption{Samples of HUST-CELL collected from various scenarios.}
\label{fig:hust_cell_expamples}
\end{figure*}

Our proposed HUST-CELL complexity goes over and above previous datasets in four distinct aspects. First, we provide 30 categories of documents with more than 4k documents, 2 times larger than the existing English and Chinese datasets including SROIE (973), CORD (1,000), EATEN (1,900), FUNSD (199), XFUNSD (1,393), and EPHOIE (1,494). Second, HUST-CELL contains 400+ diverse keys and values. Third, HUST-CELL covers complex keys more challenging than others, for instance, nested keys, fine-grained key-value pairs, multi-line keys/values, long-tailed key-value pairs, as shown in Figure~\ref{fig:hust_cell_expamples}. Current state-of-the-art Key Information Extraction (KIE) techniques~\cite{yu2021pick,zhang2020trie,xu2020layoutlm,Cao2022GMNGM,Cao2022QuerydrivenGN} fail to deal with such situations that are essential for a robust KIE system in the real world. Fourth, our dataset comprises real-world documents reflecting real-life diversity of content and the complexity of the background, e.g. different fonts, noise, blur, seal.

In this regard, under the consideration of the importance and huge application value of KIE, we propose to set up track 1 competition on complex entity linking and labeling. 

Our proposed HUST-CELL were collected from public websites and cover a variety of scenarios, e.g., receipt, certificate, and license of various industries. The language of the documents is mainly Chinese, along with a small portion of English. The number of images collected for each specific scenario varies, ranging from 10 to 300, with a long-tail distribution, which can avoid introducing any bias towards specific real application scenarios. Due to the complexity of the data source, the diversity of this dataset can be guaranteed. To be able to use publicly, this data is collected from open websites, and we delete images that contain private information for privacy protection. Some examples are shown in Figure~\ref{fig:hust_cell_expamples}.

The dataset is split into the training set and test set. The training set consists of 2,000 images, which will be available to the participants along with OCR and KIE annotations. The test set consists of 2,000 images, whose OCR annotations and KIE annotations will not be released, but the online evaluation server~\footnote{\url{https://rrc.cvc.uab.es/?ch=21\&com=mymethods\&task=1}} will remain available for future usage of this benchmark.

\subsection{Track 2: Baidu-FEST}
Our proposed Baidu-FEST benchmark comes from the practical scenarios, mainly including finance, insurance, logistics, customs inspection, and other fields. Different applications have different requirements for text fields of interest. In addition, the data collection methods in different scenarios may be affected by different cameras and environments, thus the benchmark is relatively rich and challenging. 

Specifically, the benchmark contains about 11 kinds of synthetic business documents for training, and 10 types of real visually-rich document images for testing. The format of documents major consists of cards, receipts, and forms. Each type of document provides about 60 images. 

\begin{figure*}[ht]
 \centering
 \includegraphics[width=0.8\textwidth]{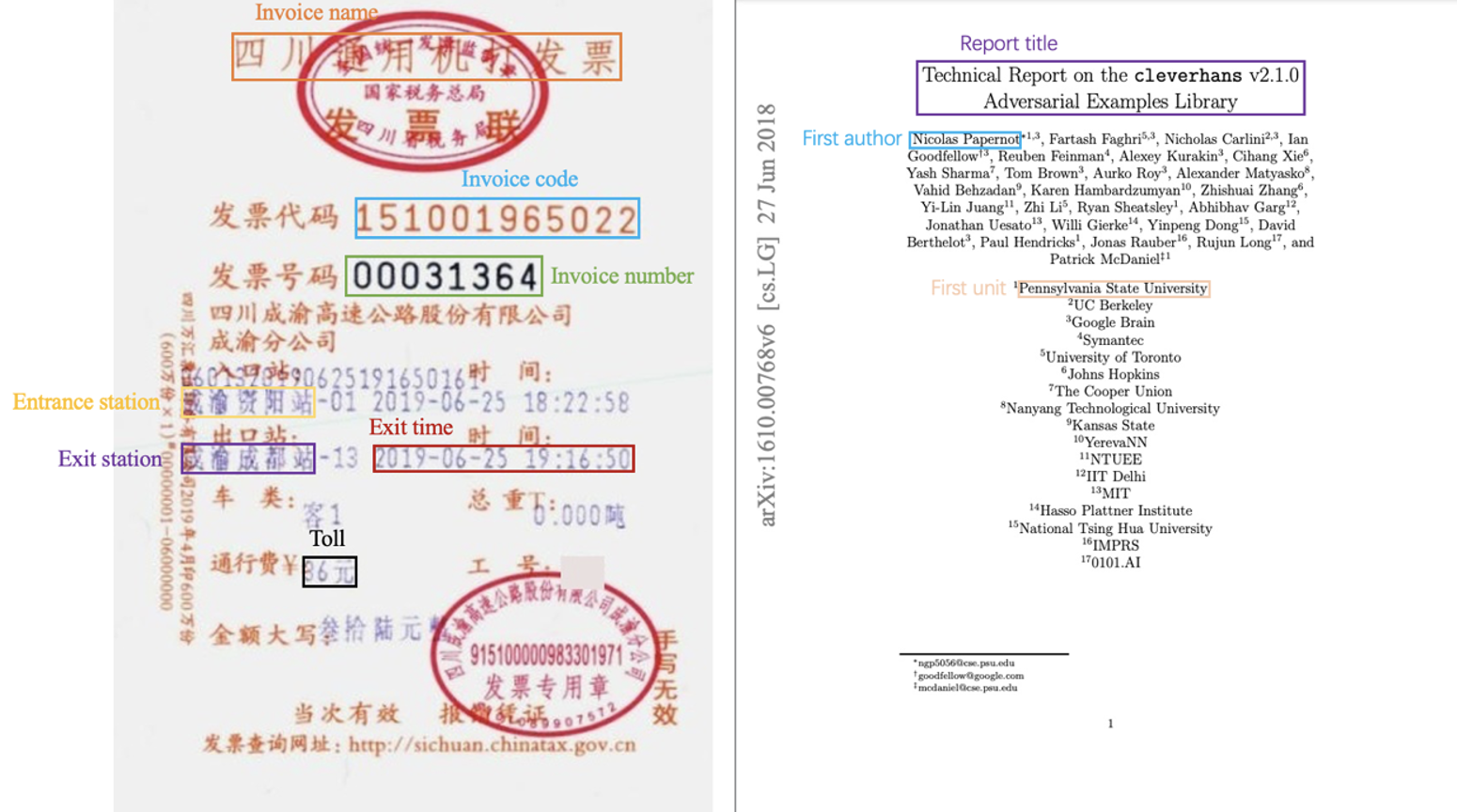}
\caption{Some visually-rich document samples of Baidu-FEST.}
\label{fig:samples}
\end{figure*}

Each image in the dataset is annotated with text-field bounding boxes (bbox) and the transcript, entity caption, entity id of each text bbox. Locations are annotated as rectangles with four vertices, which are in clockwise order starting from the top. 
Some examples of images and the corresponding annotations are shown in Figure~\ref{fig:samples}.

\section{Competition Tasks and Evaluation Protocols}
Our proposed competition has two tracks totaling four main tasks.

\subsection{Track 1: HUST-CELL}

\subsubsection{Task-1: E2E Complex Entity Linking} 
\begin{itemize}
    \item \textbf{Task Description}: This task aims to extract key-value pairs (entity linking) from given images only, then save the key-value pairs of each image into a JSON file. For the train set, both KIE annotation files for training and human-checked OCR annotation files are provided. So the OCR annotation is clean and can be used as the ground truth of the OCR task. The test set of Task 1 will only provide images without any annotation including OCR and KIE. It requires the method to accomplish both OCR and KIE tasks in an end-to-end manner.

\end{itemize}

\subsubsection{Task-2: E2E Complex Entity Labeling} 
\begin{itemize}
    \item \textbf{Task Description}: The end-to-end complex entity labeling is to extract texts of a number of predefined key text fields from given images (entity labeling), and save the texts for each image in a JSON file with required format. Task 2 has 13 predefined entities. For the train set, both KIE annotation files for training and human-checked OCR annotation files are provided. So the OCR annotation is clean and can be used as the ground truth of the OCR task. The test set of Task 2 will only provide images without any annotation including OCR and KIE. It requires the method to accomplish both OCR and KIE tasks in an end-to-end manner.
\end{itemize}
 
\subsection{Track 2: Baidu-FEST}

\subsubsection{Task-3: E2E Zero-shot Structured Text Extraction} 
\begin{itemize}
    \item \textbf{Task Description}: The zero-shot structured text extraction is to extract texts of a number of key fields from given images, and save the texts for each image in a JSON file with required format. Different from Task2, there is no intersection between the scenarios of the provided training-set and the scenarios of the provided test-set. Of course, the training data consists of the real data provided by Track 1 and  the synthetic data generated officially. The caption\_en and caption\_ch in GT can be used as prompt to assist extraction but it is not allowed to be modified.
\end{itemize}

\subsubsection{Task-4: E2E Few-shot Structured Text Extraction}
\begin{itemize}
    \item \textbf{Task Description}: The few-shot structured text extraction is to extract texts of a number of key fields from given images, and save the texts for each image in a JSON file with required format. Different from Task-3, the localization information and transcript will be provided, but the total number of the provided training-set will no more than five images for each scenario of the provided test-set. The caption\_en and caption\_ch in GT can be used as prompt to assist extraction but it is not allowed to be modified.
  
\end{itemize}

\subsection{Evaluation Protocol}

\subsubsection{Task 1 Evaluation}

For Task 1, the evaluation metrics include two parts:

\textbf{Normalized edit distance.} For each predicted kv-pair (key-value pair), if it matched with GT kv-pair in the given image, the normalized edit distance (NED) between the predicted kv-pair s1 and ground-truth kv-pair s2 will be calculated as following:

{\small
\begin{equation}
N E D(s 1, s 2)=\left(\frac{ed\left(s 1 \_k, s 2 \_k\right)}{\max \left(\operatorname{len}\left(s 1 \_k\right), \operatorname{l e n}\left(s 2 \_k\right)\right)}+\frac{ ed\left(s 1 \_v, s 2_{\_} v\right)}{\max \left(\operatorname{len}\left(s 1_{-} v\right), \operatorname{l e n}\left(s 2 \_v\right)\right)}\right) / 2
\end{equation}
}
{\small
\begin{equation}
\operatorname{score} 1=1-\sum_{i=1}^n \frac{N E D\left(s_{i 1}, s_{i 2}\right)}{n}
\label{eq:score1}
\end{equation}
}
\noindent where n denotes the number of matched kv-pairs (both the edit distance of key and value are larger than a threshold simultaneously. $ed()$ denotes the edit distance function. The calculated details refer to the following Matching Protocol.). s1\_k/s2\_k, s1\_v/s2\_v indicate the content of key and value of the kv-pair s1/s2, respectively. Note that for predicted kv-pairs that do not matched in the GT of the given image, the edit distance will be calculated between predicted kv-pairs and empty string.

\textbf{Matching Protocol}: Given the predicted kv-pair s1 and ground-truth kv-pair s2. The matching protocol is calculated as following:

{\small
\begin{equation}
    \operatorname{Match}(\mathrm{s} 1, \mathrm{~s} 2)=
\begin{cases}
True,& ed(s1\_k, s2\_k) \leq th\_k~and~ed(s1\_v, s2\_v) \leq th\_v  \\
False,& \text{other}
\end{cases}
\end{equation}
}
{\small
\begin{equation}
th\_k = max(factor\_k * min(len(s1\_k),len(s2\_k)),0)
\end{equation}
}
{\small
\begin{equation}
th\_v = max(factor\_v * min(len(s1\_v),len(s2\_v)),0)
\end{equation}
}
\noindent where $ed()$ denotes the edit distance function. the factor\_k and factor\_v are set to 0.15, 0.15, respectively.

\textbf{F-score.} Considering all the predicted kv-pairs and all GT kv-pairs, the F-score will be calculated as following:
{\small

\begin{equation}
\text { Precision }=\frac{N 3}{N 2}
\end{equation}
\begin{equation}
\text { Recall }=\frac{N 3}{N 1}
\end{equation}
\begin{equation}
\text { score}2=\frac{2 * \text { Precision } * \text { Recall }}{\text { Precision }+ \text { Recall }}
\label{eq:score2}
\end{equation}
}
\noindent where N1 denotes the number of kv-pairs that exists in the given image, N2 denotes the number of predicted kv-pairs, N3 denotes the number of perfectly matched kv-pairs (both the edit distance of key and value are larger than a threshold simultaneously. Specifically, the factor\_k and factor\_v in matching protocol are set to 0.). The final score is the weighted score of score1 and score2:
\begin{equation}
\text { score }=0.5 * \text { score}1+0.5 * \text { score}2
\end{equation}

The final weighted score will be used as submission ranking purpose for Task 1.

\subsubsection{Task 2-4 Evaluation}

For Task 2, Task 3, and Task 4, the evaluation metrics include two parts:

\textbf{Normalized edit distance.} For each predefined key text field, if it exists in the given image, the normalized edit distance (NED) between predicted text s1 and ground-truth text s2 will be calculated as following:
{\small
\begin{equation}
\begin{gathered}
N E D(s 1, s 2)=\frac{\text { edit\_distance }(s 1, s 2)}{\max (\text { length }(s 1), \operatorname{length}(s 2))} \\
\end{gathered}
\end{equation}
}
\noindent where n denotes the number of perfectly matched key text fields (both entity\_id and text are predicted correctly). Note that for predicted key text fields that do not exist in the given image, the edit distance will be calculated between predicted text and empty string. Then the $score1$ can be calculated by Eq.~\ref{eq:score1}.

\textbf{F-score.} Considering all the predicted key text fields and all the predefined key text fields, the $score2$ (F-score) can also be calculated as Eq.~\ref{eq:score2}. In this scenarios, N1 denotes the number of key text fields that exists in the given image, N2 denotes the number of predicted key text fields, N3 denotes the number of perfectly matched key text fields (both entity\_id and text are predicted correctly). 
The final score is the weighted score of score1 and score2:
\begin{equation}
\text { score }=0.8 * \text { score}1+0.2 * \text { score}2
\end{equation}

The final weighted score will be be used as submission ranking purpose for Task 2, Task 3, and Task 4.

\section{Submissions and Results}
The competition attracted 50 participants and 117 submissions from academia and industry, including 19 participants and 53 submissions for Task 1, 16 participants and 38 submissions for Task 2, 7 participants and 15 submissions for Task 3 and 8 participants and 11 submissions for Task 4, which demonstrated significant interest in this challenging task.

After the submission deadlines, we collected all submissions and evaluate their performance through automated process with scripts developed by the RRC web team. No feedback was given to the participants during the submission process. If participants have multiple submissions, we pick the last submission made before the final submission deadline for ranking. The winners are determined for each task based on the score achieved by the corresponding primary metric. The complete leaderboard is available on the online website~\footnote{\url{https://rrc.cvc.uab.es/?ch=21\&com=evaluation\&task=1}} for all tasks. However, due to limited space, the results table presented below showcases a maximum of 10 top performers.

\subsection{Task 1 Performance and Ranking}
The result for Task 1 is presented on Table~\ref{tab:task1_results}.

\begin{table*}[htbp]
\centering
\setlength\tabcolsep{2.0pt}
\caption{Task-1: E2E Complex Entity Linking Results. $^*$ means the reproducible script has been submitted by participants and verified by organizers.}
\scalebox{0.62}{
\begin{tabular}{lllllll}
\hline
Rank & Method Name & Team Members & Insititute & Score1(NED) & Score2(F-score) & Score(Total) \\
\hline
1    & Super\_KVer$^*$ & \makecell[l]{Lele Xie, \\Zuming Huang, Boqian Xia,\\ Yu Wang, Yadong Li,\\  Hongbin Wang, Jingdong Chen}& Ant Group& 49.93\%     & 62.97\%         & 56.45\%      \\
\hline
2    & \makecell[l]{End-to-end \\document relationship \\extraction}                              & Huiyan Wu, Pengfei Li, Can Li                                                                                                                                                  & \makecell[l]{University of \\Chinese Academy \\of Sciences}                                 & 43.55\%     & 57.90\%         & 50.73\%      \\
\hline
3    & sample-3$^*$                                                                 & \makecell[l]{Zhenrong Zhang,\\ Lei Jiang, Youhui Guo,\\ Jianshu Zhang, Jun Du}                                                                                                                 & \makecell[l]{University of \\Science and Technology \\of China,\\ iFLYTEK AI Research} & 42.52\%     & 56.68\%         & 49.60\%      \\
\hline
4    & \makecell[l]{Pre-trained model \\based fullpipe \\pair extraction \\(opti\_v3, no inf\_aug)$^*$} &\makecell[l] {Zening Lin, Teng Li,\\ Wenhui Liao, Jiapeng Wang,\\ Songxuan Lai, Lianwen Jin}                                                                                                    & \makecell[l]{South China \\University of Technology;\\ Huawei Cloud}                        & 42.17\%     & 55.63\%         & 48.90\%      \\
\hline
5    & Meituan OCR V4$^*$                                                           & \makecell[l]{Jianqiang Liu, Kai Zhou,\\ Chen Duan, Shuaishuai Chang,\\ Ran Wei, Shan Guo}                                                                                                & Meituan                                                                   & 41.10\%     & 54.55\%         & 47.83\%      \\
\hline
6    & submit-trainall                                                          & hsy                                                                                                                                                                          & -                                                                         & 40.65\%     & 52.98\%         & 46.82\%      \\
\hline
7    & f2                                                                       & Zhi Zhang                                                                                                                                                                    & cocopark                                                                  & 41.07\%     & 50.82\%         & 45.94\%      \\
\hline
8    & \makecell[l]{LayoutLM \& STrucText\\ Based Method}                                       & Wumin Hui, Mei Jiang                                                                                                                                                         & PKU \& BUPT                                                               & 33.09\%     & 45.92\%         & 39.51\%      \\
\hline
9    & Layoutlmv3                                                               & \makecell[l]{Li Jie,\\ Wang Wei, Li Songtao, \\Yang Yunxin, Chen Pengyu,\\ Zhou Danya, Li Chao, Hu Shiyu,\\ Zhang Yuqi, Xu Min, Zhao Yiru,\\ Zhang Bin, Zhang Ruixue, \\Wang Di, Wang Hui, Xiang Dong} & SPDB LAB                                                                  & 29.81\%     & 41.45\%         & 35.63\%      \\
\hline
10   & Data Relation2                                                                   & -                                                                                                                                                                            & -                                                                         & 23.26\%     & 35.07\%         & 29.16\%  \\
\hline
\end{tabular}}
\label{tab:task1_results}
\end{table*}

The methods used by the top 3 submissions for Task 1 are
presented below.

\textbf{1st ranking method.}  ``Ant Group'' team  apply an ensemble of both discriminative and generative models. The former is a multimodal method which utilizes text, layout and image, and they train this model with two different sequence lengths, 2048 and 512 respectively. The texts and boxes are generated by independent OCR models. The latter model is an end-to-end method which directly generates K-V pairs for an input image.

\textbf{2nd ranking method.} ``University of Chinese Academy of Sciences'' team realized end-to-end information extraction through OCR, NER and RE technologies. Text information extracted by OCR and image information are jointly transmitted to NER to identify key and value entities. RE module extracts entity pair relationships through multi-classification. The training dataset is Hust-Cell.

\textbf{3rd ranking method.} ``University of Science and Technology of China (USTC), iFLYTEK AI Research'' firstly perform key-value-background triplet classification for each OCR bounding box using a PretrainedLM called GraphDoc~\cite{zhang2022multimodal} which utilizes text, layout, and visual information simultaneously. Then they use a detection model (DBNet~\cite{liao2020real}) to detect all the table cells in input images and split images into table-images and non-table images. For table images, they merge ocr boxes into table cells and then group all the left and top keys for each value table cell as its corresponding key content. For non-table images (including all text in non-table images and text outside tabel cells in table images), they directly use a MLP model to predict all keys for each value box.

\subsection{Task 2 Performance and Ranking}
The result for Task 2 is presented on Table~\ref{tab:task2_results}.

\begin{table*}[htbp]
\centering
\setlength\tabcolsep{2.0pt}
\caption{Task-2: E2E Complex Entity Labeling Results. $^*$ means the reproducible script has been submitted by participants and verified by organizers.}
\scalebox{0.62
}{
\begin{tabular}{lllllll}
\hline
Rank & Method Name & Team Members & Insititute & Score1(NED) & Score2(F-score) & Score(Total) \\
\hline

1    & LayoutLMV3\&StrucText$^*$                                                                      & \makecell[l]{Minhui Wu, Mei Jiang,\\ Chen Li, Jing Lv,\\ Qingxiang Lin, Fan Yang}                                                                                                & TencentOCR                                                                & 57.78\%     & 55.32\%         & 57.29\%      \\
\hline
2    & sample-3$^*$                                                                                   & \makecell[l]{Zhenrong Zhang,\\ Lei Jiang, Youhui Guo,\\ Jianshu Zhang, Jun Du}                                                                                                   & \makecell[l]{University of Science \\and Technology of China,\\ iFLYTEK AI Research} & 47.15\%     & 41.91\%         & 46.10\%      \\
\hline
3    & \makecell[l]{task 1 transfer learning \\LiLT + task3 transfer \\learning LiLT + LilLT +\\ Layoutlmv3 ensemble$^*$} &\makecell[l] {Hengguang Zhou, Zeyin Lin,\\ Xingjian Zhao, Yue Zhang,\\ Dahyun Kim, Sehwan Joo,\\ Minsoo Khang, Teakgyu Hong}                                                        & Deep SE x Upstage HK                                                      & 45.70\%     & 40.20\%         & 44.60\%      \\
\hline
4    & LayoutMask-v3$^*$                                                                              & Yi Tu                                                                                                                                                          & Ant Group                                                                 & 44.79\%     & 42.53\%         & 44.34\%      \\
\hline
5    & \makecell[l]{Pre-trained model \\based entity extraction (ro)$^*$}                                             & \makecell[l]{Zening Lin, Teng Li,\\ Wenhui Liao, Jiapeng Wang, \\Songxuan Lai, Lianwen Jin}                                                                                      & \makecell[l]{South China \\University of Technology,\\ Huawei Cloud}                        & 44.98\%     & 40.06\%         & 43.99\%      \\
\hline
6    & EXO-brain for KIE                                                                          & \makecell[l]{Boqian Xia, Yu Wang,\\ Yadong Li, Zuming Huang,\\ Lele Xie, Jingdong Chen,\\ Hongbin Wang}                                                                            & Ant Group                                                                 & 44.02\%     & 39.63\%         & 43.14\%      \\
\hline
7    & \makecell[l]{multi-modal based KIE \\ through model fusion}                                                 & \makecell[l]{Jie Li, Wei Wang,\\Min Xu, Yiru Zhao,\\Bin Zhang, Pengyu Chen,\\Danya Zhou, Yuqi Zhang,\\Ruixue Zhang, Di Wang,\\Hui Wang, Chao Li,\\Shiyu Hu, Dong Xiang,\\Songtao Li, Yunxin Yang} & SPDB LAB                                                                  & 42.42\%     & 37.97\%         & 41.53\%      \\
\hline
8    & Aaaa                                                                                       & Li Rihong, Zheng Bowen                                                                                                                                         & \makecell[l]{Shenzhen Runnable \\Information Technology\\ Co.,Ltd}                          & 42.03\%     & 37.14\%         & 41.05\%      \\
\hline
9    & donut                                                                                      & zy                                                                                                                                                             & -                                                                         & 41.64\%     & 37.65\%         & 40.84\%      \\
\hline
10   & Ant-FinCV                                                                                  & Tao Huang, Jie Wang, Tao Xu                                                                                                                                    & Ant Group                                                                 & 41.61\%     & 35.98\%         & 40.48\%     \\
\hline
\end{tabular}}
\label{tab:task2_results}
\end{table*}

The methods used by the top 3 submissions for Task 2 are
presented below.

\textbf{1st ranking method.} ``TencentOCR'' team are mainly based on LayoutLMv3 and StrucTextv1 model architecture. All training models are finetuned on large pretrained models of LayoutLM and StrucText. During training and testing, they did some preprocessings to merge and split some badly detected boxes. Since entity label of kv-pair boxes are ignored, they used model trained on task1 images to predict kv relations of text boxes in Task 2 training/testing images. Thus they added additional 2 classes of labels (question/answer) and mapped original labels to new labels(other $\rightarrow$ question/answer) to ease the difficulty of training. Similarly, During testing, they used kv-prediction model to filter those text boxes with kv relations and used model trained on Task 2 to predict entity label of the lefted boxes. In addition, they combined predicted results of different models based on scores and rules and did some postprocessings to merge texts with same entity label and generated final output.

\textbf{2nd ranking method.} ``University of Science and Technology of China (USTC), iFLYTEK AI Research'' team uses the GraphDoc~\cite{zhang2022multimodal} to perform bounding box classification, which utilizes text, layout, and visual information simultaneously.

\textbf{3rd ranking method.} ``Deep SE x Upstage HK'' team, for the OCR, uses a cascade approach where the pipeline is broken up into text detection and text recognition. For text detection, they use the CRAFT architecture with the backbone changed to EfficientUNet-b3. For text recognition, they use the ParSeq architecture with the visual feature extractor changed to SwinV2. Regarding the parsing models, they trained both the LiLT and LayoutLMv3 models on the Task2 dataset. For LiLT, they also conducted transfer learning on either task1 or task3 before fine-tuning on Task 2 dataset. Finally, they take an ensemble of these four models to get the final predictions.

\subsection{Task 3 Performance and Ranking}
The result for Task 3 is presented on Table~4.%

\begin{table*}[htbp!]
\centering
\footnotesize
\setlength\tabcolsep{0.5pt}
{
\caption{Task-3: E2E Zero-shot Structured Text Extraction Results. $^*$ means the reproducible script has been submitted by participants and verified by organizers.}
\scalebox{0.7}{
\begin{tabular}{lllllll}
\hline
Rank & Method Name & Team Members & Insititute & Score1(NED) & Score2(F-score) & Score(Total) \\
\hline
1    & sample-1$^*$     & \makecell[l]{Zhenrong Zhang, Lei Jiang, \\Youhui Guo, Jianshu Zhang,\\ Jun Du}                                  & \makecell[l]{University of Science \\and Technology of China,\\ iFLYTEK AI Research} & 82.07\%     & 65.27\%         & 78.71\%      \\
\hline
2    & LayoutLMv3$^*$   & \makecell[l]{Minhui Wu, Mei Jiang, Chen Li,\\ Jing Lv, Huiwen Shi}                                            & TencentOCR & 80.01\%     & 66.71\%         & 77.35\%      \\
\hline
3    & KIE-Brain3$^*$   &\makecell[l]{ Boqian Xia, Yu Wang,\\ Yadong Li, Ruyi Zhao, \\Zuming Huang, Lele Xie, \\Jingdong Chen, Hongbin Wang} & Ant Group& 74.90\%     & 57.59\%         & 71.44\%      \\
\hline
4    & zero-shot-qa & - & - & 74.24\%     & 56.81\%         & 70.75\%      \\
\hline
5    & task3-2      & chengl & CMSS& 65.52\%     & 50.85\%         & 62.59\% \\
\hline

\end{tabular}
}
}
\label{tab:task_three_results}
\end{table*}

The methods used by the top 3 submissions for Task 3 are
presented below.

\textbf{1st ranking method.} ``University of Science and Technology of China (USTC), iFLYTEK AI Research'' team first use OCR models (DBNet-det + SVTR-rec) to get each bounding box coordinate and it's text content of input images, then sort boxes with manual rules, and concatenate all text content to a string accroding to the box sequence. Result string of step 1 is fed into a seq2seq model to directly predict target text content, which consists of 8 open-source bert models, including chinese-roberta, chinese-lilt-roberta, chinese-pert, chinese-lilt-pert, chinese-lert, chinese-lilt-lert, chinese-macbert and chinese-lilt-macbert, and these models are trained using the UniLM~\footnote{https://github.com/microsoft/unilm/blob/master/s2s-ft/} toolbox. Data augmentation including random text replacing and erasing, box random scaling and shifting is used. As for english doc images, they directly use DocPrompt~\footnote{https://github.com/PaddlePaddle/PaddleNLP/blob/develop/model\_zoo/ernie-layout} outputs as final result

\textbf{2nd ranking method.} ``TencentOCR'' team's method is based on LayoutLMv3 and StrucTextv1 model architecture. All training models are finetuned on large pretrained models of LayoutLM and StrucText. During training and testing, we did some preprocessings to merge and split some badly detected boxes.  They used models trained on task 1 images to predict key-value pairs in test images and models trained on task 2 images to predict entity labels(title, date, etc) for text boxes. Besides, they applied rule based post processing methods  and assembled results of different models to generate final outputs.

\textbf{3rd ranking method.} ``Ant Group'' team apply  an ensemble of multi-task end-to-end information extraction models. The document question answering task and the document information extraction task are jointly realized, and the model performance is improved. At the same time, this solution is an end-to-end information extraction method and does not rely on external OCR.

\subsection{Task 4 Performance and Ranking}
The result for Task 4 is presented on Table~\ref{tab:task4_results}.

\begin{table*}[htbp]
\centering
\setlength\tabcolsep{2.0pt}
\caption{Task-4: E2E Few-shot Structured Text Extraction Results. $^*$ means the reproducible script has been submitted by participants and verified by organizers.}
\scalebox{0.68}{
\begin{tabular}{lllllll}
\hline
Rank & Method Name & Team Members & Insititute & Score1(NED) & Score2(F-score) & Score(Total) \\
\hline
1    & \makecell[l]{LayoutLMv3\&\\StrucText}$^*$         & \makecell[l]{Mei Jiang, Minhui Wu,\\ Chen Li, Jing Lv,\\ Haoxi Li, \\Lifu Wang, Sicong Liu}                                                 & TencentOCR                                                                & 87.14\%     & 73.59\%         & 84.43\%      \\
\hline
2    & sample-1$^*$                      & \makecell[l]{Zhenrong Zhang,\\ Lei Jiang, Youhui Guo, \\Jianshu Zhang, Jun Du}                                                            & \makecell[l]{University of Science\\ and Technology of China,\\ iFLYTEK AI Research} & 85.24\%     & 69.68\%         & 82.13\%      \\
\hline
3    & task4-base                    & chengl                                                                                                                  & CMSS                                                                      & 78.57\%     & 60.21\%         & 74.90\%      \\
\hline
4    & Fewshot-brain\_v1$^*$             & \makecell[l]{Boqian Xia, Yu Wang, \\Yadong Li, Hongbin Wang}                                                                            & Ant Group                                                                 & 77.81\%     & 60.71\%         & 74.39\%      \\
\hline
5    & \makecell[l]{Dao Xianghu \\light of TianQuan} & \makecell[l]{Kai Yang, Tingmao Lin,\\ Ye Wang, Shuqiang Lin, \\Jian Xie, Bin Wang,\\ Wentao Liu, Xiaolu Ding, \\Jun Zhu, Hongyan Pan,\\ Jia Lv} & \makecell[l]{CCB Financial Technology \\Co. Ltd, China }                                  & 71.48\%     & 55.03\%         & 68.19\%      \\
\hline
6    & GRGBanking                    & \makecell[l]{Liu Kaihang, Yue Xuyao,\\Xu Tianshi, Zhang Huajun, \\Liang Tiankai} & GRGBanking                                                                 & 45.44\%     & 35.83\%         & 43.52\%\\
\hline
\end{tabular}}
\label{tab:task4_results}
\end{table*}

The methods used by the top 3 submissions for Task 4 are
presented below.

\textbf{1st ranking method.} ``TencentOCR'' team's methods are mainly based on  LayoutLMv3 and StrucTextv1 model architecture. All training models are finetuned on large pretrained models of LayoutLM and StrucText. During training and testing, they did some preprocessings to merge and split some badly detected boxes. They also trained our own ocr models  including dc-convnet based detection and ctc/eda based recognition. They applied merging methods to merge ocr results from different sources. Based on predicted results of task 3, they also introduced  self-supervising training to train segment-based classification models for folder 9 \& folder 10 to predict entity labels. Similar to task 3, they also did  rule based post processing methods  and assembled results of different models to generate final outputs.

\textbf{2nd ranking method.} ``University of Science and Technology of China (USTC), iFLYTEK AI Research'' team first use OCR models (DBNet-det + SVTR-rec) to get each bounding box coordinate and it's text content of input images, then sort boxes with manual rules, and concatenate all text content to a string accroding to the box sequence. Result string of step 1 is fed into a seq2seq model to directly predict target text content, which consists of 8 open-source bert models, including chinese-roberta, chinese-lilt-roberta, chinese-pert, chinese-lilt-pert, chinese-lert, chinese-lilt-lert, chinese-macbert and chinese-lilt-macbert, and these models are trained using the UniLM~\footnote{https://github.com/microsoft/unilm/blob/master/s2s-ft/} toolbox. Data augmentation including random text replacing and erasing, box random scaling and shifting is used. As for english doc images, they directly use DocPrompt~\footnote{https://github.com/PaddlePaddle/PaddleNLP/blob/develop/model\_zoo/ernie-layout} outputs as final result.

\textbf{3rd ranking method.} ``CMSS'' team's method firstly perform data cleaning, data synthesis, and random augmentation based on the given data. For blurry images, they utilize a variety of cascaded models to optimize the text detection results. To further improve the accuracy of text recognition algorithm, they use a neural network structure and remove special characters. Particularly, they have found that the position of the text plays a significant role in the structure, especially for structured text with key and value. To address this issue, they optimize the model parameters based on the UIE-X basic model (Unified Structure Generation for Universal Information Extraction) and optimize the data pre-processing part to enhance the relationship features between text lines. Finally, they merge the model inference results by designing rules to achieve the best possible outcome.

\section{Discussion}
The competition saw a considerable number of submissions from both academic and industrial participants, indicating a notable level of interest in the topic of complex structured text extraction. The competition focused on two tracks: Track 1 involved end-to-end complex entity linking and labeling, while Track 2 introduced new zero/few-shot scenario tasks for structured text extraction.%

The leading approaches in Track 1 of the competition utilized ensemble models that integrated multiple modalities, including text, layout, and image information. These models were enhanced by incorporating techniques such as OCR, NER, and RE. The prominence of multimodal models and ensemble techniques highlights their significance. 
Nevertheless, there exists a notable performance gap in achieving the necessary accuracy for information extraction in complex scenarios for large-scale practical applications. The observed performance gap is notable when comparing the top-performing F1 score of 55.32\% for Task 2 to the state-of-the-art end-to-end method on other datasets. Specifically, Kuang et al.~\cite{Kuang2023VisualIE} achieved an 85.87\% F1 score on the SROIE dataset and Donut~\cite{kim2022donut} achieved an 84.1\% F1 score on the CORD dataset in end-to-end evaluations.

Track 2 of the SVRD competition focused on new zero/few-shot scenario tasks for structured text extraction. Specifically, Task 3 required participants to extract structured text from images that were not present in the training set, whereas Task 4 allowed participants to perform structured text extraction using a small number of provided labeled examples. The methods used for zero-shot and few-shot learning involved various deep learning techniques, such as pretrained large multimodal model and model ensemble. The top-performing methods achieved 78.71\% and 84.43\% for Task 3 and Task 4, respectively. Despite the commendable average performance of Task 3 and Task 4 across all 10 scenarios, the individual performance analysis of the champion method reveals significantly low effectiveness in more challenging scenarios. Specifically, the champion method attained a score of 53.25\% in the Letter/Email scenario and 46.2\% in the Technical Report scenario for Task 3. In Task 4, the champion method achieved a score of 59.23\% in the Car Ticket scenario. These results underscore the need for further research to enhance the performance of these methods in structured text extraction for such challenging scenarios.

Despite the top-performing methods showcased promising results, a significant performance gap remains for challenging scenarios in structured text extraction. Advancements in developing robust techniques for structured text extraction are crucial for the progress of Document AI. The SVRD competition provided a platform for researchers in CV and NLP to collaborate and showcase their expertise. The results highlight the potential of structured text extraction for various document analysis applications, emphasizing the need for continued research.

\section{Conclusion}
We hosted the SVRD competition, focused on key information extraction from visually-rich document images, and provided new datasets, including HUST-CELL and Baidu-FEST, and designed new evaluation protocols for our tasks. The submissions indicated strong interest from academia and industry, but also revealed significant challenges in achieving high performance for complex and zero-shot scenarios. Key information extraction remains a difficult task with potential for numerous document analysis applications. Future competitions could expand on this topic with more challenging datasets and applications, attracting researchers in CV and NLP and advancing the field of Document AI.

\section{Acknowledgements} This competition is supported by the National Natural Science Foundation of China (No.62225603, No.62206103, No.62206104). The organizers thank Sergi Robles and the RRC web team for their tremendous support on the registration, submission and evaluation jobs.

\bibliographystyle{splncs04}
\bibliography{main}
\end{document}